%% file: main.tex
\documentclass[acmtog]{acmart}
\acmSubmissionID{1168}

\input{setup/package}   
\input{setup/color}
\input{setup/macro}     
\input{setup/graphicspath}

\input{setup/math}

\newcommand{\ourmethod}[1]{PAD3R}

\usepackage[ruled]{algorithm2e} 

\SetAlFnt{\small}
\SetAlCapFnt{\small}
\SetAlCapNameFnt{\small}
\SetAlCapHSkip{0pt}

\acmJournal{TOG}

\begin{document}

\title{PAD3R: Pose-Aware Dynamic 3D Reconstruction from Casual Videos}

\ccsdesc[500]{Computing methodologies~Animation}
\ccsdesc[500]{Computing methodologies~Shape modeling}
\ccsdesc[500]{Computing methodologies~Rendering}
\ccsdesc[300]{Computing methodologies~Computer vision}

\keywords{4D reconstruction and generation, animal reconstruction and tracking, novel view synthesis}

\author{Ting-Hsuan Liao}
\authornote{Co-first authors.}
\affiliation{%
\institution{University of Maryland College Park}
\country{USA}
\city{}}
\email{ting1129@umd.edu}

\author{Haowen Liu}
\authornotemark[1]
\affiliation{%
\institution{University of Maryland College Park}
\country{USA}
\city{}}
\email{hwl@umd.edu}

\author{Yiran Xu}
\affiliation{%
\institution{University of Maryland College Park}
\country{USA}
\city{}}
\email{yiranx@umd.edu}

\author{Songwei Ge}
\affiliation{%
\institution{University of Maryland College Park}
\country{USA}
\city{}}
\email{songweig@umd.edu}

\author{Gengshan Yang}
\authornote{Joint advising. 
}
\affiliation{%
}
\email{y.gengshan@gmail.com}

\author{Jia-Bin Huang}
\authornotemark[2]
\affiliation{%
\institution{University of Maryland College Park}
\country{USA}
\city{}}
\email{jbhuang@umd.edu}

\begin{teaserfigure}
\centering
\includegraphics[trim={0.8cm 0.5cm 0.7cm 0cm},clip,width=\linewidth]
{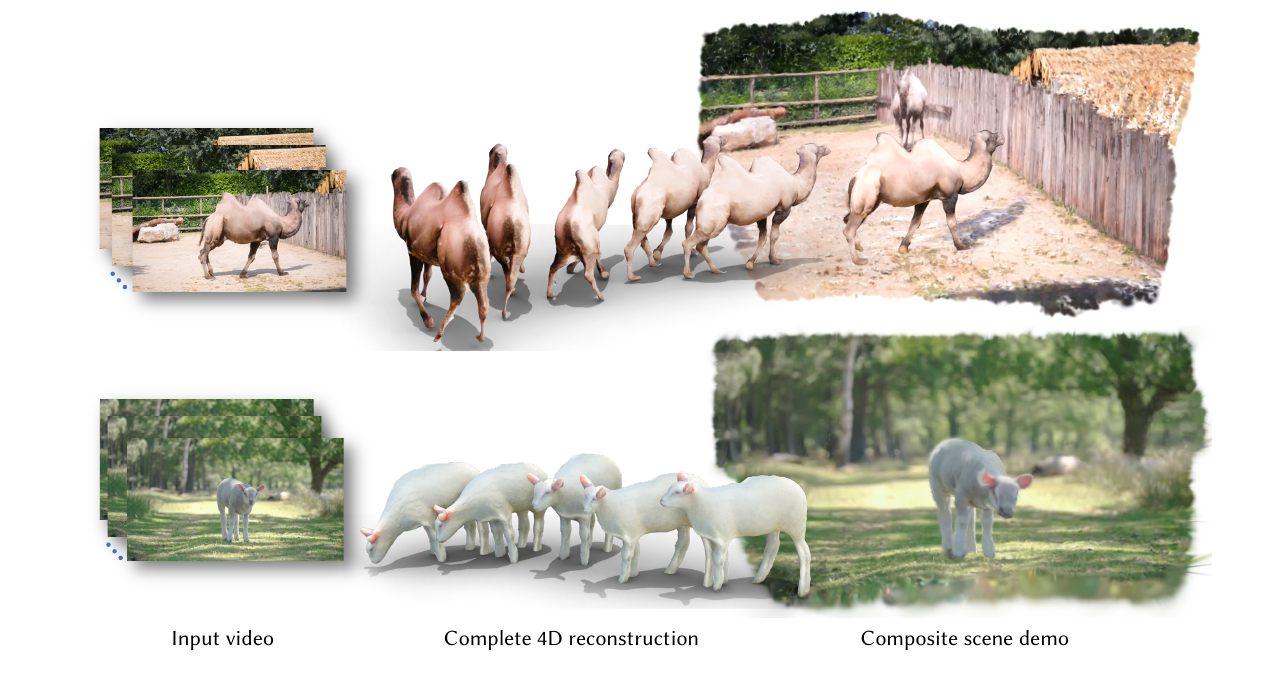}
\vspace{-0.8cm}
\caption{
\textbf{Complete 3D reconstruction of dynamic objects from a single unposed video.} 
Given a casually captured monocular video (left), our method reconstructs accurate object-centric camera poses, high-quality geometry, appearance, and non-rigid deformations over time (right). 
Leveraging the capabilities of generative models and differentiable rendering, PAD3R recovers complete, plausible 3D shape and deformations, remaining robust to significant root body, camera movement, and limited viewpoints. Images sourced from DAVIS (\textcopyright DAVIS Challenge organizers) and \textcopyright Pexels.
}
\Description{Teaser image showing, on the left, a single casual monocular video frame, on the right, our method’s reconstructed dynamic object with camera poses and deformations over time.}
\label{fig:teaser}
\end{teaserfigure}

\input{0_abstract}

\maketitle

\input{1_introduction}

\input{2_related}

\input{3_method}

\input{4_result}

\input{5_conclusions}

\input{6_acknowledgment}


\bibliographystyle{ACM-Reference-Format}
\bibliography{main}

\end{document}

%% file: setup/package.tex
\usepackage{lscape}

\usepackage{colortbl}
\usepackage{makecell}
\usepackage{multirow}

\usepackage{pifont}
\usepackage{amsfonts}
\usepackage{bm}
\usepackage{mathtools}

%% file: setup/color.tex
\colorlet{dark-blue}{blue!50!black}
\colorlet{dark-cyan}{cyan!75!black}
\colorlet{dark-purple}{purple!50!black}
\colorlet{dark-red}{red!75!black}
\colorlet{dark-green}{green!80!black}
\colorlet{dark-orange}{orange!50!black}
\colorlet{dark-gray}{black!75}
\colorlet{light-gray}{black!30}
\definecolor{nice-red}{HTML}{E41A1C}
\definecolor{nice-orange}{HTML}{FF7F00}
\definecolor{nice-yellow}{HTML}{FFC020}
\definecolor{nice-green}{HTML}{39b54a}
\definecolor{nice-blue}{HTML}{0071bc}
\definecolor{nice-purple}{HTML}{984EA3}

\colorlet{verylight-gray}{black!10}
\definecolor{LightCyan}{rgb}{0.66,0.85,0.76}



%% file: setup/macro.tex
\makeatletter
\DeclareRobustCommand\onedot{\futurelet\@let@token\@onedot}
\def\@onedot{\ifx\@let@token.\else.\null\fi\xspace}

\makeatother

\newcommand{\ignore}[1]{}   

\newcommand{\topic}[1]
{
\noindent\textbf{#1}
}

\newcolumntype{L}[1]{>{\raggedright\let\newline\\\arraybackslash\hspace{0pt}}m{#1}}
\newcolumntype{C}[1]{>{\centering\let\newline\\\arraybackslash\hspace{0pt}}m{#1}}
\newcolumntype{R}[1]{>{\raggedleft\let\newline\\\arraybackslash\hspace{0pt}}m{#1}}

\definecolor{tabfirst}{rgb}{1, 0.7, 0.7} 
\definecolor{tabsecond}{rgb}{1, 0.85, 0.7} 
\definecolor{tabthird}{rgb}{1, 1, 0.7} 

\definecolor{xmark}{rgb}{0.93, 0.39, 0.37} 
\definecolor{cmark}{rgb}{0.53, 0.66, 0.42} 

\newcommand{\cmark}{\color{cmark}{\ding{51}}}%
\newcommand{\xmark}{\color{xmark}{\ding{55}}}%

%% file: setup/graphicspath.tex
\graphicspath{{figures}, {examples}}

%% file: setup/math.tex
\def\1n{\mathbf{1}_n}
\def\0{\mathbf{0}}
\def\1{\mathbf{1}}

\def\C{{\bf C}}
\def\D{{\bf D}}

\def\F{{\bf F}}

\def\I{{\bf I}}

\def\L{{\bf L}}
\def\M{{\bf M}}

\def\P{{\bf P}}

\def\a{{\bf a}}

\def\e{{\bf e}}

\def\l{{\bf l}}
\def\m{{\bf m}}

%% file: 0_abstract.tex

\begin{abstract}
We present \ourmethod{}, a method for reconstructing deformable 3D objects from casually captured, unposed monocular videos. Unlike existing approaches, PAD3R handles long video sequences that feature substantial object deformation, large-scale camera movement, and limited view coverage, which typically challenge conventional systems. At its core, our approach trains a personalized, object-centric pose estimator, supervised by a pre-trained image-to-3D model. This guides the optimization of deformable 3D Gaussian representation. The optimization is further regularized by long-term 2D point tracking over the entire input video. By combining generative priors and differentiable rendering, PAD3R reconstructs high-fidelity, articulated 3D representations of objects in a category-agnostic way. Extensive qualitative and quantitative results show that \ourmethod{} is robust and generalizes well across challenging scenarios, highlighting its potential for dynamic scene understanding and 3D content creation. Please refer to our project page for more details: \textcolor{blue}{\href{https://pad3r.github.io/}{PAD3R.github.io}}.


\end{abstract}

%% file: 1_introduction.tex
\section{Introduction}
\label{sec:intro}
Reconstructing dynamic 3D objects from monocular videos is crucial for applications in gaming, film production, augmented and virtual reality, and robotics~\cite{yunus2024recent}.
However, this task remains challenging due to its ill-posed nature, where infinitely many 3D interpretations explain the same 2D observations. 
Classic approaches often rely heavily on specialized sensors~\cite{park2006capturing} or category-specific models~\cite{SMPL:2015}, limiting their applicability to diverse objects and scenes in real-world environments.


Recent advances in differentiable rendering techniques such as Neural Radiance Fields (NeRF) and 3D Gaussian Splatting, along with their dynamic extensions~\cite{nerfplayer, dnerf2021, nerfies, hypernerf, wu4dgaussians, dynamic3dgaussian}, have significantly improved 3D/4D reconstructions of general scenes. 
However, these approaches often require precise camera poses and dense viewpoint coverage. 
To overcome this dependency, a line of work~\cite{jiang2023consistent4d, zeng2025stag4d, ren2023dreamgaussian4d, li2024dreammesh4d, tu2023dreamo, zhang2024bags} incorporates generative priors from pre-trained 2D (or multi-view) diffusion models into 4D reconstruction pipelines using Score Distillation Sampling (SDS) and its variants. 
But such approaches are limited to videos from a static viewpoint in a synthetic setup. 
Several recent methods~\cite{xie2024sv4d, wu2024cat4dcreate4dmultiview, ren2024l4gm} finetune video diffusion models for inferring the multi-view views of dynamic 3D objects directly from monocular videos. 
Nonetheless, their effectiveness hinges on high-quality multi-view training data, limiting generalization to complex, out-of-distribution real-world videos, many of which involve large camera and object motion unseen during training. 
Hence, robust dynamic 3D reconstruction from in-the-wild monocular videos remains an open challenge.





In this paper, we introduce PAD3R, a method that tackles the challenge of learning a deformable 3D representation of a target object from a single casually captured unposed video.
Our core idea is to leverage both the generative diffusion prior (for recovering a static 3D representation) and differentiable rendering (for jointly estimating object-centric camera poses and time-varying deformation). 
Given a video, we first use an image-to-3D method~\cite{liu2023zero} to obtain a static canonical 3D model from a keyframe. 
We then train a \emph{personalized} object-centric camera pose estimator, by fine-tuning DINO-v2~\cite{oquab2023dinov2}, using renderings of the generated 3D model sampled from random viewpoints. 
This pose initialization is critical for the subsequent dynamic reconstruction stage.
We initialize a hybrid 3D Gaussian representation using the canonical geometry and predicted poses.
Optimizing the temporal deformation model solely using photometric reconstruction loss is difficult due to large object deformation, self-occlusion, and limited view coverage. 
We regularize the training with motion cues from a persistent point tracker~\cite{karaev24cotracker3}, augmented with a proposed multi-chunk strategy aggregating occluded signals across frames.
We validate the proposed method through extensive experiments on diverse real and synthetic video datasets. 
Our approach consistently outperforms state-of-the-art methods in both qualitative and quantitative evaluations. 
Ablation studies further validate the importance of our design choices in improving reconstruction accuracy and temporal coherence.

Our contributions are as follows.
\begin{itemize}
\item We leverage an image-to-3D model to learn object-centric camera poses, providing crucial initialization for the subsequent 4D reconstruction stage.
\item We introduce a multi-chunk strategy that effectively uses long-term motion cues from 2d point tracking to regularize object deformation.
\item We evaluate our method on both synthetic datasets, including Consistent4D and Artemis, as well as  challenging real-world videos. \ourmethod{} demonstrates state-of-the art 4D reconstruction results and robustness to real world camera motion. We will release our code to support and foster future research.
\end{itemize}

%% file: 2_related.tex
\section{Related Work}
\label{sec:related}


\noindent{\bf Dynamic Scene Reconstruction.}
Dynamic scene reconstruction aims to recover a dynamic 3D scene from videos, supporting both viewpoint- and time-varying rendering.
Building on the success of Neural Radiance Fields (NeRFs)~\citep{mildenhall2020nerf}, several works~\cite{dnerf2021, nerfies, hypernerf, gao2021dynerf} have extended NeRFs to dynamic settings.
However, due to the volumetric rendering nature of these methods, they often require time-consuming optimization, even with efficiency-focused advancements~\cite{kplanes, hexplane}.
Recently, 3D Gaussian Splatting (3DGS)~\cite{3dgaussian} introduces a significantly faster rendering pipeline, making high-quality, real-time dynamic scene rendering possible.
A series of work~\cite{dynamic3dgaussian,wu4dgaussians,yang2024deformable,yang2023real,duan20244d} extend 3DGS by introducing an additional deformation field.
However, these methods struggle to model complex non-rigid deformations or sparse views.
Although regularization~\cite{guedon2023sugar,wang2024shape, lei2024moscadynamicgaussianfusion,yang2022banmo} is employed, they still struggle with reconstructing a monocular video, resulting in suboptimal results.
More recent works~\cite{wang2024shape,wang2025gflow,stearns2024dynamic,lei2024moscadynamicgaussianfusion,qingmingmodgs} tackle reconstruction from a single monocular video, but primarily focus on scenes.
These methods demonstrate limited view ranges in novel view rendering.
In contrast, our method focuses on \emph{full 360° reconstruction} of dynamic 3D objects with complex deformations from a single monocular video.

\input{figures/fig_overview}

\noindent{\bf 4D Reconstruction with Generative Priors.}
Inspired by the success of generative modeling~\cite{ho2020denoising, song2020denoising}, recent works~\cite{jiang2023consistent4d, zeng2025stag4d, ren2023dreamgaussian4d, pan2024efficient4d, li2024dreammesh4d, tu2023dreamo, wang2023animatabledreamer, zhang2024bags, zhao2023animate124, yin20234dgen} have explored incorporating generative priors into 4D reconstruction pipelines to reduce reliance on densely-captured multi-view data.
Score Distillation Sampling (SDS) and its variants~\cite{poole2022dreamfusion, wang2023prolificdreamer, yi2023gaussiandreamer} are commonly adopted to supervise reconstruction from unseen views.
However, these methods typically assume fixed camera poses and root-body object motions, severely limiting their applicability to real-world videos with complex camera trajectories and dynamic object motion.
In contrast, we propose a more robust approach to handle such challenging scenarios.


\noindent{\bf 4D Reconstruction with Feed-Forward Models.}
Recent methods~\cite{xie2024sv4d, wu2024cat4dcreate4dmultiview, ren2024l4gm, park2025zero4d} directly train video diffusion or feed-forward models for spatio-temporally consistent 4D generation from a monocular video, without relying on test-time optimization.
However, their performance heavily depends on the quality and diversity of multi-view video data used during training.
As a result, these methods often struggle to generalize to out-of-distribution real-world videos, particularly those with complex object and camera motions not seen in the training datasets~\cite{ren2024l4gm, deitke2023objaverse, xie2024sv4d, wu2024cat4dcreate4dmultiview}.
In this paper, we introduce a robust method, PAD3R, for addressing such challenges and demonstrate superior performance over feed-forward baselines.




%% file: figures/fig_overview.tex
\begin{figure*}[t]    
    \centering
    \includegraphics[trim={0.3cm 0.2cm 0cm 0cm},clip,width=0.95\textwidth]{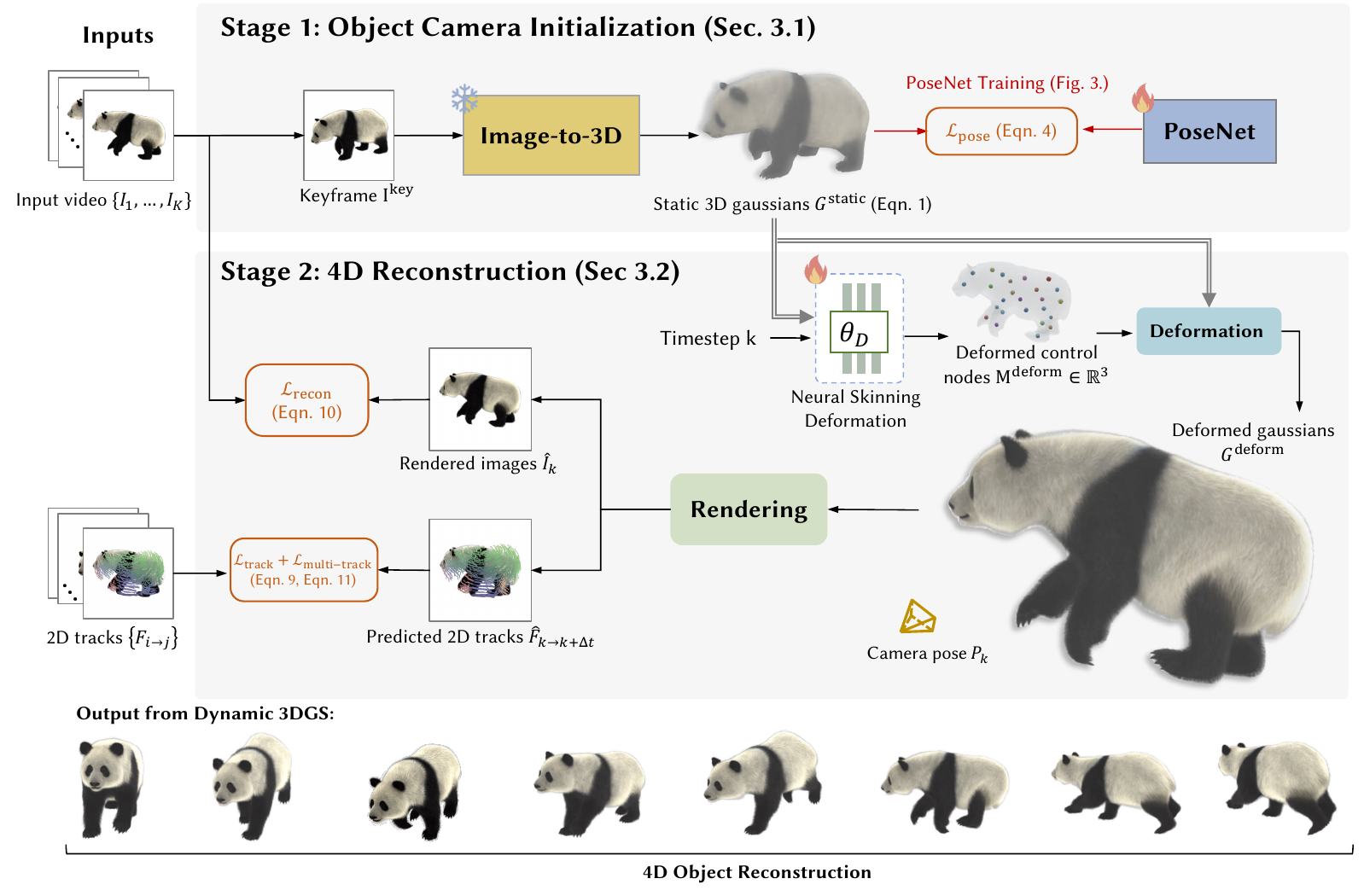}
    \caption{\textbf{Method overview.} 
    Our method consists of two main stages. 
    In the first stage, we select a frame from the video sequence as the canonical frame (or keyframe), and use an image-to-3D model~\cite{liu2023zero} to obtain a static 3D Gaussian $\hat{G}$. 
    We then render $\hat{G}$ from a set of randomly sampled camera poses 
    to fine-tune a lightweight image-to-pose estimator, PoseNet, using DINO-v2 backbone. 
    The full training scheme is illustrated in Figure.~\ref{fig:posenet}. Then, we use the camera pose estimator to initialize the camera pose of every input video frame and optimize a deformable 3D object model. Images sourced from Artemis (\textcopyright Luo et al., 2022).
    }
    \Description{Method overview.}
    \label{fig:overview}
\end{figure*}

%% file: 3_method.tex


\def\D{\mathcal{D}}  
\def\I{\mathcal{I}}  
\def\m{\mathcal{M}}  
\def\M{\mathcal{M}}  
\def\C{\mathcal{C}}  
\def\F{\mathcal{F}}  
\def\a{\alpha}       
\def\P{P}            
\def\l{l}  
\newcommand{\nv}[1]{\tilde{#1}} 

\def\e{\epsilon}

\def\L{\mathcal{L}} 

\def\ng{\hspace{-0.1mm}}
\def\neg{\hspace{-0.2mm}}
\def\pos{\hspace{0.2mm}}

\section{Method}
\label{sec:method}
Given a monocular video sequence $\{I_1, I_2, \dots, {I}_{K}\}$ containing \textit{K} RGB images that depict a dynamic, articulated object (such as a person or animal), our objective is to reconstruct its complete time-varying 3D geometry. 
This task is inherently ill-posed and we solve it in two stages: 1) camera pose initialization and 2) dynamic Gaussian splatting reconstruction. 
An overview of our approach is in Figure~\ref{fig:overview}.

\subsection{Object-centric Camera Pose Initialization}
\label{subsec:camera}

A key ambiguity in dynamic 3D reconstruction from in-the-wild videos comes from the entanglement of camera and object motions.
As we focus on reconstructing the \emph{dynamic object}, we need to explicitly model the camera poses with respect to the dynamic object (not to the scene).
Note that this is a more challenging case than estimating camera poses with respect to the static scene due to the lack of rigidity constraints. 

Instead of solving both camera and object motions simultaneously through optimization,
we first estimate the camera pose of each input frame, \emph{relative to the object’s coordinate system}.
These estimates serve as initialization for the subsequent joint optimization stage. While our method targets non-rigid objects, we assume their overall shape and appearance remain relatively consistent throughout the video. 
This assumption enables reliable pose estimation by training a personalized pose prediction model for each object instance, using only a single-frame 3D reconstruction as supervision, illustrated in Figure~\ref{fig:posenet}.

We begin by selecting a keyframe $I^\mathrm{key}$ from the video and generating a 3D representation conditioned on this image. \begingroup
The keyframe is selected such that the object appears in a common, well-posed configuration (e.g., the side view of a standing animal), which helps avoid inaccuracies in generation due to extreme or unusual poses.
\endgroup

To overcome the geometric limitations of na\"ive 3DGS in geometric modeling, we adopt SuGaR~\cite{guedon2023sugar}, a hybrid Gaussian representation. 
SuGaR enhances surface modeling by introducing a regularization term that aligns Gaussians with mesh surfaces, enabling more accurate geometry and mesh-based operations such as deformation.



We denote the SuGaR model as $G^\mathrm{static} = \{\mathcal{V}, \mathcal{F}, \hat{G}\} $, where $\mathcal{V}$ and $\mathcal{F}$ are the mesh vertices and faces, respectively, and $\hat{G}$ is the set of 3D Gaussians anchored to mesh surfaces.
Each face $f \in \mathcal{F}$ is associated with a fixed number of Gaussians, whose positions are parameterized using barycentric coordinates.
The resulting set of Gaussians $\hat{G}$ is defined as:


\begin{equation}
   \hat{G} = \{\mathit{\mu_i}, \mathit{\Sigma_i,} \mathit{c_i}, \mathit{\alpha_i}\}_{i=1}^{N},
\end{equation}
where $\mathit{\mu_i}$ is the mean, $\mathit{\Sigma_i}$ is a positive semi-definite covariance matrix, $\mathit{c_i}$ is the view-dependent color parametrized using spherical harmonics (SH), and $\mathit{\alpha_i}$ is the opacity value.

Following the approach in~\cite{li2024dreammesh4d}, we begin by generating a coarse mesh using an off-the-shelf image-to-3D method~\cite{liu2023zero}. 
To enrich surface detail, we attach six flat Gaussians to each triangle face of the mesh.
We then optimize the SuGaR model with photometric reconstruction loss $\mathcal{L}_{photo}$ on the reference view and Score Distillation Sampling (SDS) loss on unseen views:

\begin{equation}
    \mathcal{L}_{\text{3D}} = \lambda_{\text{photo}}\mathcal{L}_{\text{photo}} + \lambda_{\text{sds}} \mathcal{L}_{\text{sds}}, 
\end{equation}
where
\begin{equation}
\mathcal{L}_{\text{sds}} = \mathbb{E}_{\mathbf{x} \sim \mathcal{N}(0, I), t \sim \mathcal{U}(0, T)} \left[ w(t) \left\| \boldsymbol{\epsilon}_{\theta}(\mathbf{x}_t, t) - \boldsymbol{\epsilon} \right\|_2^2 \right].
\end{equation}
Here, $\mathbf{x}_t$ denotes the noisy latent at timestep $t$, $\boldsymbol{\epsilon}_{\theta}$ is the predicted noise from the score network, $\boldsymbol{\epsilon}$ is the ground truth noise, and $w(t)$ is a time-dependent weighting function.

Using the optimized Gaussian model---though any other 3D representation could be used---we synthesize images $C_{l}$ by rendering with randomly sampled camera poses $\pi_l = \left( \mathbf{R}_l,\mathbf{T}_l \right)$. 
We train the pose estimation model $E$ on these rendered images, augmented with color jitter, random masking, and rotations. 
The pose estimation model $E$ uses a pre-trained DINOv2~\cite{oquab2023dinov2} backbone with a multi-layer perceptron (MLP) regression head to predict the 6-DoF camera pose in the object coordinate system, by minimizing the pose loss $\mathcal{L}_{\text{pose}}$:

\input{figures/fig_posenet}

\begin{equation}
\mathcal{L}_{\text{pose}} = \lambda_{\text{rot}}\mathcal{L}_{\text{rot}} + \lambda_{\text{trans}}\mathcal{L}_{\text{trans}} + \lambda_{\text{unc}}\mathcal{L}_{\text{unc}},
\end{equation}
where 
$\mathcal{L}_{\text{trans}}$ and $\mathcal{L}_{\text{rot}}$ denote the translational and rotational components respectively:

\begin{equation}
     \mathcal{L}_{\text{trans}} = \| \mathbf{T} - \hat{\mathbf{T}} \|^2_2, \quad \mathcal{L}_{\text{rot}} = \arccos((\text{Tr}(\mathbf{R}\hat{\mathbf{R}}^T)-1)/2),
\end{equation}

where $\pi = \left( \mathbf{R},\mathbf{T} \right)$ denotes the ground-truth camera pose consisting of rotation $\mathbf{R}$ and translation $\mathbf{T}$, and $\hat{\pi} = \left( \hat{\mathbf{R}},\hat{\mathbf{T}} \right)$ is the predicted poses from the network.
In addition, we incorporate an uncertainty loss $\mathcal{L}_{\text{unc}}$ to account for the confidence of the pose predictions~\cite{kendall2017uncertainties}, encouraging the predicted uncertainty $\sigma$ to reflect the actual pose error magnitude:
%
%
\begin{equation}
\mathcal{L}_{\text{unc}} =  (\sigma - (\mathcal{L}_{\text{rot}} + \mathcal{L}_{\text{trans}}))^2,
\end{equation}


Using the trained pose estimator, we initialize the camera pose ${P}_{k}$ for each input frame ${I}_{k}$ in the video sequence as initialization for subsequent optimizations.

\begin{figure}[t]
    \centering
    \includegraphics[trim={0.3cm 0cm 0.3cm 0cm},clip,width=\linewidth]{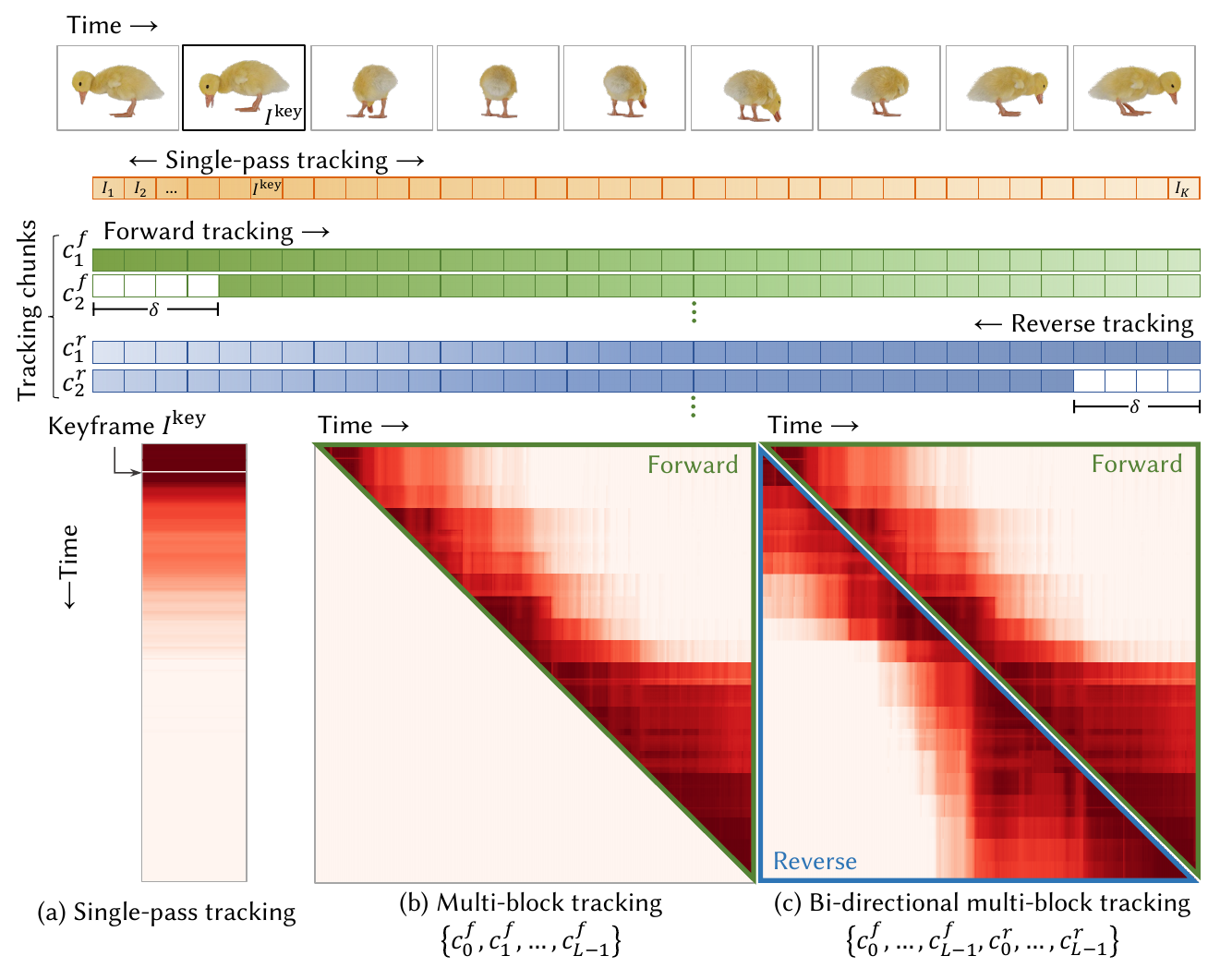}
    \caption{
    \textbf{Co-visibility of tracked points across time.} 
    Given a video sequence $\{I_1, I_2, \dots, I_K\}$, we extract dense correspondences using a 2D tracking model~\cite{karaev24cotracker3}. 
    Each colored block represents the frame range tracked from a specific starting frame, with a temporal stride of $\delta = 4$ in this illustration. 
    (a) \textbf{Single-pass tracking starting from the canonical frame $I^{\text{key}}$} provides strong guidance in the canonical space but has limited coverage in distant frames due to occlusion or narrow field of view. 
    (b) \textbf{Multi-block tracking} (we show forward tracking $\{c^f_0, c^f_1, \dots\, c^f_{L-1} \}$ here) improves coverage by dividing the video into chunks and tracking forward only, but lacks symmetric temporal consistency. 
    (c) \textbf{Our proposed bi-directional multi-block tracking} strategy combines forward $\{c^f_i\}$ and reverse $\{c^r_i\}$ tracking across overlapping chunks, yielding dense, bidirectional correspondences between all frame pairs. Images sourced from Artemis (\textcopyright Luo et al., 2022).}
    \Description{Co-visibility figure.}
    \label{fig:track-ill}
\end{figure}

\subsection{Dynamic Gaussian Splats Reconstruction}

\topic{Neural Skinning Deformation Model}
To model dynamic motion, we uniformly sample a set of control nodes $M$ anchored to mesh vertices in the static SuGaR model $G^\mathrm{static}$.
The deformation model $\theta_D$, conditioned on the video timestep $k$, predicts per-frame deformations of these nodes.

We adopt the hybrid deformation framework from ~\cite{li2024dreammesh4d}, where each vertex deformation is computed by blending transformations from neighboring control nodes using a combination of Linear Blend Skinning (LBS)~\cite{lbs} and Dual Quaternion Skinning (DQS)~\cite{dqs}. 
Each control node carries rotation, shear, and translation, and a learned rigidity score that modulates its influence. 
The final deformation at each vertex is computed by interpolating between LBS and DQS results, weighted by the aggregated local rigidity from neighboring nodes.


\topic{Dense tracking supervision}
While photometric losses can provide appearance-based supervision from input frames, but they offer limited guidance on dynamic object motion. 
To address this, we incorporate dense 2D tracking ~\cite{karaev24cotracker3} which captures temporal correspondences and motion dynamics in image space.
This enables spatially and temporally coherent supervision for 4D reconstruction from monocular video.
Given a sequence of $K$ frames, the tracker predicts a set of tracked 2D points across the sequence, including pixel locations and visibility masks. 
We denote the tracked 2D coordinates as $x_k^d \in \mathbb{R}^2$ and visibility as $v_k^d \in \{0,1\}$, where $k \in [0, K)$ and $d \in [0, D)$ indexes tracked points.

To supervise the canonical 3D model over time, we construct single-pass point trajectories anchored at the selected keyframe.
Starting from the selected keyframe $I^\mathrm{key}$, we run the tracker in both forward ($I^\mathrm{key} \rightarrow K$) and backward ($I^\mathrm{key} \rightarrow 0$) through the video, resulting in long-range, temporally consistent trajectories aligned with the canonical view.

Using the reconstructed 3D mesh at the keyframe $I^\mathrm{key}$, we lift each 2D point $x_\mathrm{key}^d$ into a 3D point $X^d \in \mathbb{R}^3$ on the surface via ray-mesh intersection. 
These 3D anchors establish correspondence between 2D tracks and the canonical 3D space.

However, a single frame can only cover a small field of view, leaving many pixels in the distant frames to be untracked. 
To address this, we apply bi-directional multi-block tracking by partitioning the video into $L$ overlapping chunks to increase track density and supervision coverage:

\begin{align}
\text{forward:} & \quad [k_0 : K],\ [k_1 : K],\ \dots,\ [k_{L-1} : K]  \\
\text{reverse:} & \quad [K - k_0 : 0],\ [K - k_1 : 0],\ \dots,\ [K - k_{L-1} : 0], 
\end{align}
where $k_i = i \cdot  \delta $, for $k=0,1,...,L-1$, and $\delta$ is the temporal stride.

Each chunk tracks points from its start frame $k_i$ to the end of the video, ensuring that for any frame pair $(i, j)$, at least one chunk covers both frames.
To select the optimal chunk for each pair $(i, j)$, we evaluate the number of commonly visible tracks across all chunks and chose the one with the highest overlap. 
We illustrate the proposed sampling strategy in Figure~\ref{fig:track-ill}.

For for each tracked point $x_\mathrm{key}^d$ with the known 3D position $ X^d$ from single-pass tracking, we find the corresponding Gaussian splatting points $g^d$ in the canonical model via nearest-neighbor search in 3D.
At each timestep $k$, we project these Gaussian points $g^d$ to obtain the predicted 2D location $\hat{x}_t^d$. 
We compute a tracking loss as 
\begin{equation}
\mathcal{L}_{\text{track}} = \sum_{k, d} v_k^d \cdot | \hat{x}_k^d - x_k^d |
\end{equation}
where $v_k^d$ is a visibility mask that excludes the invisible points.

For bi-directional multi-block tracking, we compute the 2D tracks between any two frames $(i, j)$ as:
\begin{equation}
\mathbf{F}_{i \rightarrow j}^d = x_j^d - x_i^d,
\end{equation}
where $d \in \mathcal{D}_{i,j}$ is visible in both frames $i$ and $j$.

For every two frames, we compute its motion in 3D space across frames and project it to 2D track $\hat{\mathbf{F}}_{i,j}^d$ via rasterization. This prediction is supervised against the tracked flow with a masked L2 loss:
\begin{equation}
\mathcal{L}_{\text{multi-track}} = \sum_{i<j,\ d} m_{i,j}^d \cdot \| \hat{\mathbf{F}}_{i \rightarrow j}^d - \mathbf{F}_{i \rightarrow j}^d \|^2_2,
\end{equation}
where $m_{i,j}^d$ indicates visibility of point $d$ in both frames.

This dual strategy delivers robust, spatially dense supervision, enabling accurate and consistent reconstruction of dynamic geometry even under challenging visibility and motion conditions.

\topic{Optimization}
The overall objective in the reference view reconstruction stage combines several losses.
In addition to a photometric loss that supervises appearance consistency, we apply an as-rigid-as-possible (ARAP) regularization term~\cite{10.5555/1281991.1282006} to promote locally consistent deformation of the hybrid Gaussian surface.

The ARAP loss penalizes deviations from locally rigid transformations between neighboring vertex pairs and is defined as:
\begin{equation}
\mathcal{L}_{\text{ARAP}} = \sum_{\mathbf{v} \in \mathcal{V}} \sum_{\mathbf{v}_n \in \mathcal{N}(\mathbf{v})} \omega_n(\mathbf{v}) \left\| (\tilde{\mathbf{v}} - \tilde{\mathbf{v}}_n) - R_{\mathbf{v}}(\mathbf{v} - \mathbf{v}_n) \right\|_2^2.
\end{equation}
Here, $\mathcal{V}$ is the set of mesh vertices, $\mathcal{N}(\mathbf{v})$ denotes the 1-ring neighbors of vertex $\mathbf{v}$. The variables $\omega_n(\mathbf{v})$ is the cotangent weight associated with neighbor $\mathbf{v}_n$. The variables $\tilde{\mathbf{v}}$ and $\tilde{\mathbf{v}}_n$ represent deformed vertex positions, and $R_{\mathbf{v}}$ is the locally estimated rotation at vertex $\mathbf{v}$.

The full loss jointly optimizes both the deformation network and the pose network and is formulated as:
\begin{equation}
\mathcal{L}_{\text{4D}} = \lambda_{\text{rgb}}\mathcal{L}_{\text{rgb}} + \lambda_{\text{track}} \mathcal{L}_{\text{track}} + \lambda_{\text{multi}} \mathcal{L}_{\text{multi-track}} + \lambda_{\text{arap}} \mathcal{L}_{\text{ARAP}},
\end{equation}
where each weight $\lambda$ controls the contributions of its corresponding term. The photometric loss 
$\mathcal{L}_{\text{rgb}} = \| \hat{I} - I \|_2^2$ measures the mean squared error between the rendered image $\hat{I}$ and the ground truth input frame $I$.
Additionally, we optimize a per-frame delta camera pose parameterized by an MLP simultaneously.

%% file: figures/fig_posenet.tex
\begin{figure}[t]
    \centering
    \includegraphics[trim={0.1 0.2cm 0.1 0cm},clip,width=\linewidth]{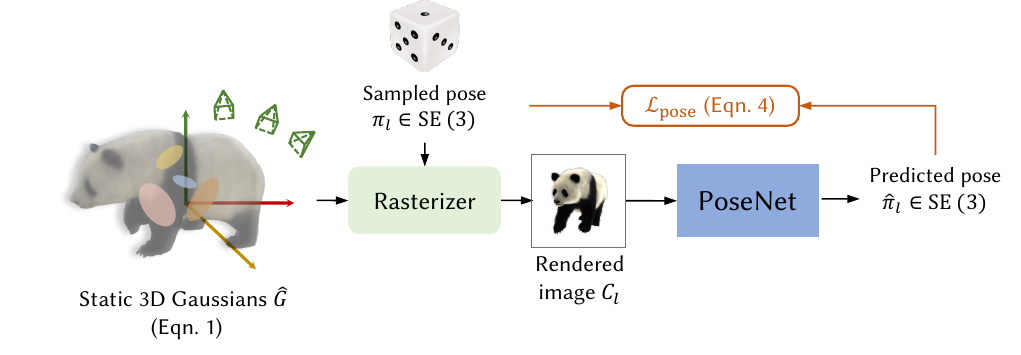}
    \caption{\textbf{Training personalized PoseNet.} To train the pose estimator network $E$, we first obtain a static Gaussian model of the selected canonical frame $I^\mathrm{key}$. At each training iteration, we render images $C_{l}$ on-the-fly from the model with randomly sampled camera poses. These images are used to train the network to predict 6-DoF camera poses, parameterized by translations and quaternion rotations.}
    \Description{Posenet overview.}
    \label{fig:posenet}
\end{figure}

%% file: 4_result.tex
\section{Experiments}
\label{sec:result}

\subsection{Dataset}
We quantitatively evaluate PAD3R on the Consistent4D~\cite{jiang2023consistent4d} dataset, which includes seven multi-view videos, each with 32 frames. 
For each sequence, a single-view video is used as input, while the remaining four views serve as evaluation references.
Since Consistent4D features \emph{static} cameras only, we introduce a new benchmark based on the Artemis dataset~~\cite{10.1145/3528223.3530086}.
The dataset contains multi-view videos of animals with varying camera motion and viewpoint coverage. 
To create this benchmark, we sample camera views that collectively span a target angular range around the animal, assembling these views into a single turntable-style video for evaluation.

For qualitative evaluation, we use a collection of real-world videos sourced from the internet, the DAVIS dataset~\cite{pont2017davis}, and from BANMo~\cite{yang2022banmo}. 
These clips are captured in casual, in-the-wild environments, exhibiting diverse and complex camera and object motions.
Each video ranges from 40 to 300 frames in length, providing rich temporal dynamics for assessment.

\subsection{Implementation Details}


We train the pose estimation network for 4,000 iterations using the Adam optimizer with a learning rate of $5 \times 10^{-4}$, $\beta_1 = 0.9$, $\beta_2 = 0.999$. 
We train the deformation model for 2,000 iterations using the AdamW optimizer with $\beta_1 = 0.9$ and $\beta_2 = 0.99$ for the reference view reconstruction stage. 
The Posenet training takes 1 hour, and the optimization of the deformation model takes 1.5 hours on a single NVIDIA A6000 GPU.


\subsection{Baselines}
We compare PAD3R with video-to-4D reconstruction methods, including STAG4D~\cite{zeng2025stag4d}, L4GM~\cite{ren2024l4gm}, DreamMesh4D (DM4D)~\cite{li2024dreammesh4d} and BANMo~\cite{yang2022banmo}. STAG4D, L4GM and DreamMesh4D do not account for dynamic camera motion, operating under the assumption of a static camera.



\input{tables/quantative_comp_c4d}

\subsection{Quantitative Results}
We follow the evaluation protocol from Consistent4D~\cite{jiang2023consistent4d}, reporting on perceptual similarity using LPIPS~\cite{lpips}, CLIP~\cite{radford2021clip} for image similarity, and FVD~\cite{unterthiner2018fvd} metric to assess video temporal coherence between the ground truth and rendered novel view images. 
Results are summarized in Table~\ref{tab:quant_c4d}. 
Under the static camera setting, PAD3R outperforms prior approaches, including  STAG4D~\cite{zeng2025stag4d}, DreamMesh4D~\cite{li2024dreammesh4d}, and L4GM~\cite{ren2024l4gm}, achieving the best scores across all metrics.
Furthermore, in a dynamic camera setting (i.e., where we estimate the camera poses by assuming they are unknown), our method remains competitive, surpassing BANMo~\cite{yang2022banmo}, which also explicitly models camera motion. 

For the Artemis benchmark, we select camera views spanning various viewing angles around each object and compile them into turntable video sequences of 20 to 300 frames. 
Evaluation is conducted using 3 novel views per sequence, each with corresponding ground truth data for quantitative comparison.


We follow the same evaluation protocol as in Consistent4D~\cite{jiang2023consistent4d}, reporting LPIPS, CLIP and FVD to assess perceptual quality and temporal coherence of the novel views. 
Quantitative results are presented in Table~\ref{tab:quant_artmeis}. 
PAD3R achieves state-of-the-art performance across all metrics, outperforming prior approaches. 
These results demonstrate the effectiveness of our model in accurately capturing both camera motion trajectories and the complex non-rigid dynamics of objects.

\input{tables/quantative_comp}

\begin{figure*}[t]    
    \centering
    \includegraphics[trim={0 0cm 0 0.cm},clip,width=1.0\textwidth]{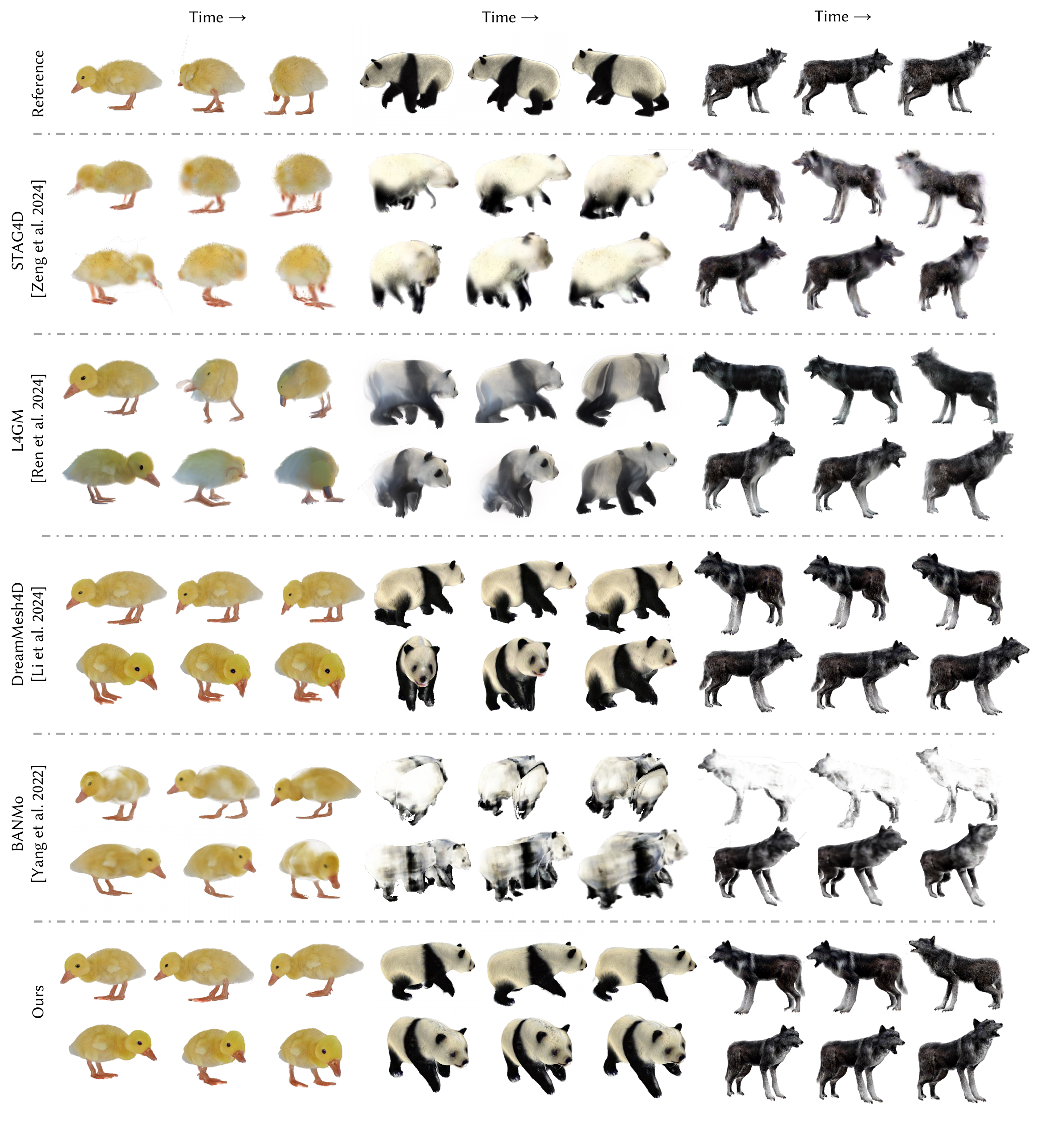}
    \caption{Comparison of our method against baseline methods on the Artemis dataset. Our method can generate novel views of better quality while remaining faithful to the reference views. DreamMesh4D~\cite{li2024dreammesh4d} suffers from distortions in the mouth and chest regions for the duck sequence, and has the wrong leg geometry for the panda sequence, while our reconstructions do not suffer from those problems. Images sourced from Artemis (\textcopyright Luo et al., 2022).}
    \Description{Showing comparison on Artemis dataset.}
    \label{fig:artemis_qualitative}
\end{figure*}

\begin{figure}[t]    
    \centering
    \includegraphics[trim={0 .3cm 0 0.1cm},clip,width=0.8\linewidth]{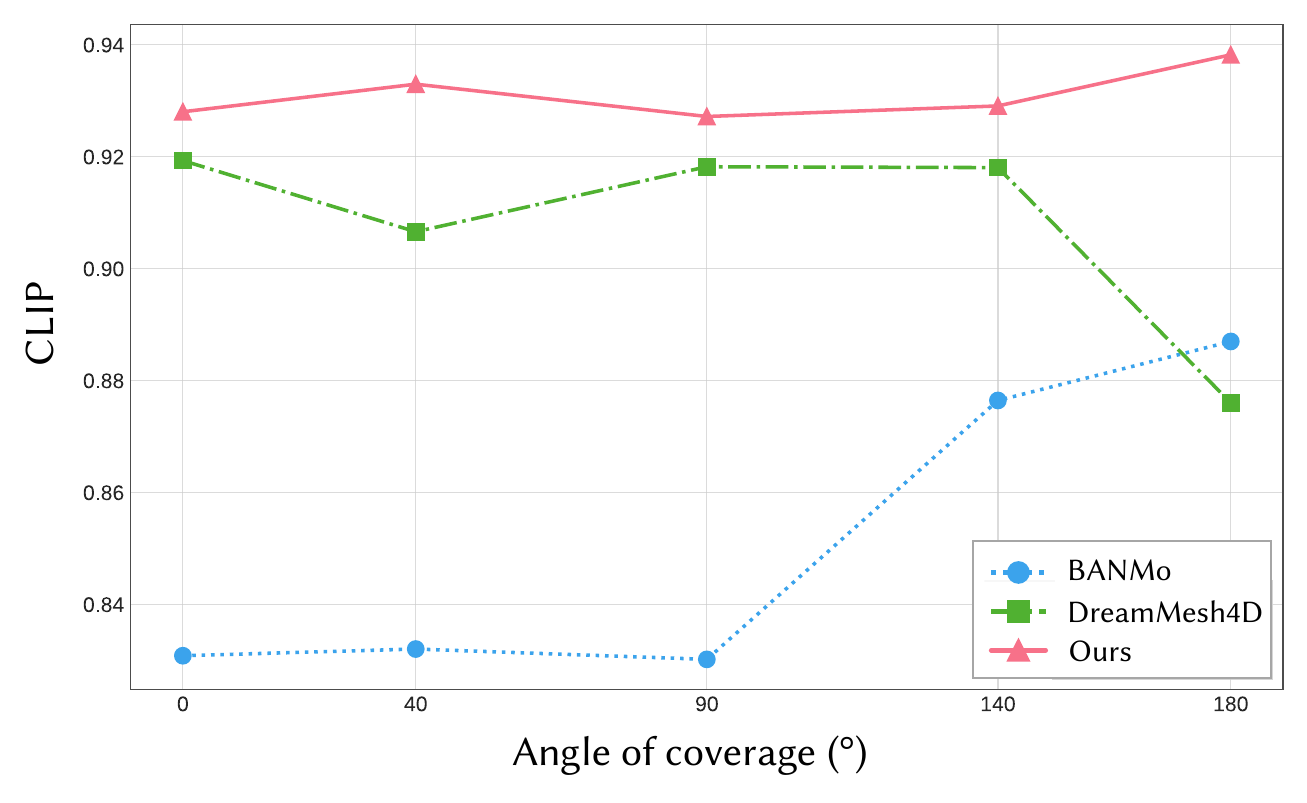}
    \caption{
    \textbf{Reconstruction quality \textit{v.s.} input view coverage.} PAD3R consistently achieves high performance across different amounts of view coverage. In contrast, DreamMesh4D~\cite{li2024dreammesh4d} degrades under large camera motion, while BANMo~\cite{yang2022banmo} performs better with broader view coverage but struggles in limited view cases.}
    \Description{Showing performance on different view coverage.}
    \label{fig:coverage}
\end{figure}

To further assess PAD3R’s capacity to capture camera movements, we analyze how input view coverage affects reconstruction quality. 
Specifically, we evaluate on five input sequences, each providing a different extent of viewpoint variation,  0\textdegree (single-view), 40\textdegree, 90\textdegree, 140\textdegree and 180\textdegree. Figure~\ref{fig:coverage} shows the reconstruction performance against input view coverage of our method against the baseline method DreamMesh4D~\cite{li2024dreammesh4d}, the current state-of-the-art method that assumes a static camera, and BANMo~\cite{yang2022banmo}, which explicitly models camera motion.

As illustrated in Figure~\ref{fig:coverage}, PAD3R maintains consistently high reconstruction quality across varying view coverage angles. 
In contrast, due to its static camera assumption, DreamMesh4D exhibits a steady decline in performance as the range of viewpoints expands.
Conversely, BANMo~\cite{yang2022banmo} shows improved results with broader view coverage, but performs poorly under single-view or narrow-view settings. 
These results underscore the robustness and adaptability of our approach in handling both constrained and dynamic camera motions, enabling accurate reconstructions under diverse conditions.

\input{figures/fig_ablation}



\subsection{Qualitative Results}
We showcase qualitative results comparing our method against several baseline approaches, including STAG4D~\cite{zeng2025stag4d}, DreamMesh4D~\cite{li2024dreammesh4d}, L4GM~\cite{ren2024l4gm}, and BANMo~\cite{yang2022banmo}. 
These evaluations span both controlled synthetic scenes from the Artemis dataset~\cite{10.1145/3528223.3530086} and casually captured, in-the-wild videos. 

\begin{figure*}[t]    
    \centering
    \includegraphics[trim={0cm 2cm 0cm 0cm},clip,width=1.0\textwidth]
    {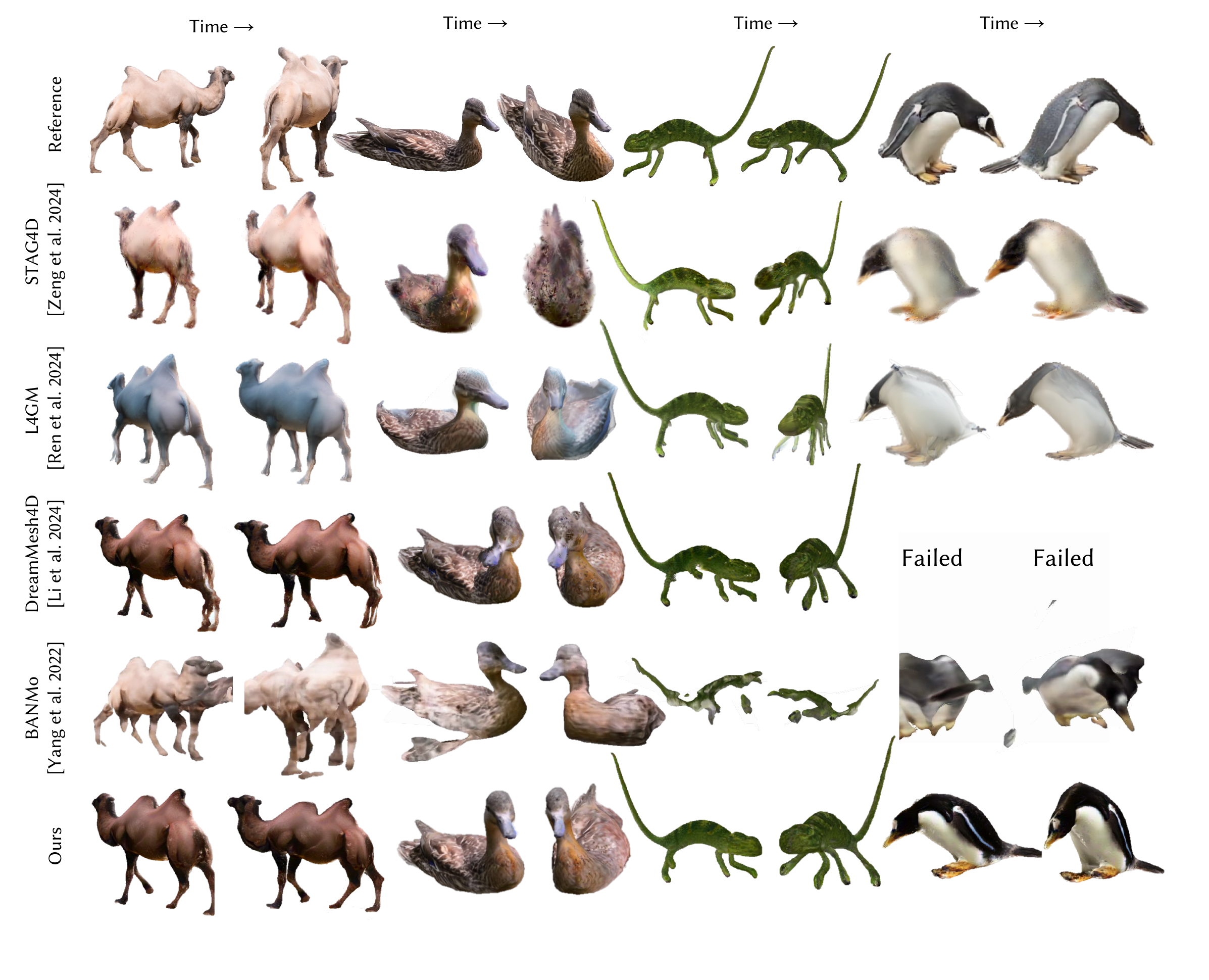}
    \caption{Qualitative comparison against baseline methods on challenging in-the-wild videos with large camera and object root body motion. Ours is able to achieve consistently high-quality reconstruction while baseline methods fail. Images sourced from DAVIS (\textcopyright DAVIS Challenge organizers) and \textcopyright Pexels.}
    \Description{Showing qualitative comparison on in-the-wild videos.}
    \label{fig:qualitative-comparison}
\end{figure*}

Figure~\ref{fig:artemis_qualitative} showcases qualitative comparisons on the Artemis dataset~\cite{10.1145/3528223.3530086}, illustrating PAD3R’s ability to recover high-fidelity textures and accurate geometry under complex object and camera motions. 
In Figure~\ref{fig:qualitative-comparison}, we extend the evaluation to real-world scenarios. 
Our method generalizes well to real-world videos, delivering complete, coherent, and stable 3D reconstructions across diverse motions.
\begingroup
The supplementary material includes additional results and dynamic visualizations showcasing reconstructions and novel view renderings.
\endgroup

\input{tables/quantative_comp_ablat}

\subsection{Ablation Studies}
We conduct an ablation study to evaluate the impact of core components in our method: object-centric camera modeling, PoseNet-based pose initialization, and the two proposed tracking objectives: $\mathcal{L}_{track}$ and $\mathcal{L}_{multi-track}$.
Quantitative results are reported in Table~\ref{tab:quant_ablt}, and qualitative comparisons are shown in Figure~\ref{fig:ablation}. 
We select our final model configuration based on the best overall performance across metrics.

The ablation demonstrates that all components are critical in achieving accurate and detailed 4D reconstruction. 
PoseNet initialization enhances camera pose estimation, resulting in improved reconstruction. 
The $\mathcal{L}_{track}$ loss captures fine-grained motion, particularly around articulated limbs. 
Incorporating $\mathcal{L}_{multi-track}$ further boosts
temporal consistency by aligning object and camera dynamics.
Together, these elements significantly improve both quantitative performance and visual fidelity.

\input{tables/quantative_alt_pose}

\begingroup
\topic{Impact of Pose Initialization.}
We further conduct experiments comparing our personalized PoseNet against state-of-the-art camera pose estimators, MegaSaM~\cite{megasam} and VGGT~\cite{vggt}, to evaluate the effect of initialization quality.
As reported in Table~\ref{tab:quant_alt}, using PoseNet predictions consistently achieves the best reconstruction quality without camera refinement.
In contrast, off-the-shelf models struggle to estimate accurate object-centric poses due to the lack of static correspondences, leading to degraded reconstruction.
Even after enabling camera refinement, our model still achieves the strongest performance when initialized with PoseNet. The experiments demonstrate the importance of reliable object-centric pose estimation.
These results underscore the need for personalized pose estimation to achieve robust and temporally coherent 4D reconstruction.
\endgroup

\begin{figure*}[t]    
    \centering
    \includegraphics[trim={2.8cm 15.5cm 3.8cm 6.5cm},clip,width=0.9\textwidth]
    {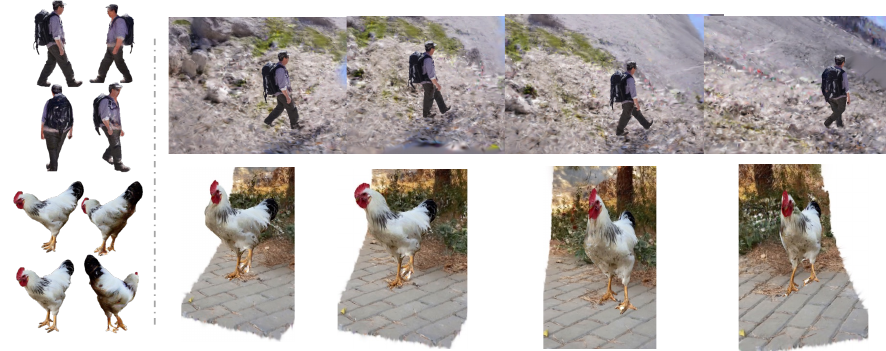}
    \caption{Composing our object-level 4D reconstructions (left) with static background Gaussians produces a full dynamic scene reconstruction (right). Images sourced from DAVIS (\textcopyright DAVIS Challenge organizers) and \textcopyright Pexels.}
    \Description{Showing compositional scene.}
    \label{fig:demo}
\end{figure*}

\subsection{Limitations}
While PAD3R has demonstrated improved performance over the state-of-the-art on dynamic 3D object reconstruction from videos, it has several limitations.
First, since our method employs per-video optimization, it is not suitable for applications that require real-time or rapid processing.
Second, while off-the-shelf image-to-3D models have shown great progress, they sometimes produce erroneous 3D shapes and overly smooth textures. 
This, in turn, may adversely affect the training of the pose estimation model. 
This can be potentially mitigated by jointly learning the canonical 3D shape and appearance model (together with camera poses and deformation) directly from the video. 
However, our initial attempts found it challenging due to the ambiguity. 
Third, the 2D tracking models may be inaccurate, particularly when rapid motion or heavy occlusion occurs.
Also, no constraints can be derived from unseen viewpoints. 
To address this, we believe that using motion priors (e.g., from video diffusion models) is a promising future direction.

%% file: tables/quantative_comp_c4d.tex
\begin{table}[t]
    \centering
    \caption{\textbf{Quantitative Results on Consistent4D Dataset.} Consistent4D Dataset includes videos captured from a single viewpoint and is evaluated across four novel views. The training schemes of STAG4D~\cite{zeng2025stag4d} and DreamMesh4D (DM4D) ~\cite{ren2024l4gm} align with data that lacks camera or scene motion, which contrasts with our method's capability to model dynamic camera movements. For this evaluation dataset, PAD3R achieves comparable results against STAG4D~\cite{zeng2025stag4d} and DreamMesh4D~\cite{ren2024l4gm} and outperforms BANMo~\cite{yang2022banmo} that model camera dynamics. $\colorbox[RGB]{255,179,179}{\:}$: best, $\colorbox[RGB]{255,217,179}{\:}$: second-best, $\colorbox[RGB]{255,255,179}{\:}$: third-best.   
}
    \resizebox{0.48\textwidth}{!}{%
    \begin{tabular}{lcccc}
        \toprule
        \thead{\textbf{Methods}} & \thead{\textbf{Camera} \\ \textbf{modeling}} & \thead{\textbf{LPIPS$~\downarrow$}} & \thead{\textbf{FVD$~\downarrow$}} & \thead{\textbf{CLIP $~\uparrow$}} \\
        \midrule
        STAG4D~\cite{zeng2025stag4d} &\xmark  & \cellcolor{tabthird}0.134 & 1015.57 &  0.917  \\
        L4GM~\cite{ren2024l4gm} & \xmark & 0.152 & 874.49 & 0.921 \\
        DM4D~\cite{li2024dreammesh4d} & \xmark &  \cellcolor{tabsecond}0.128          & \cellcolor{tabthird}688.84 & \cellcolor{tabthird}0.936    \\
        BANMo~\cite{yang2022banmo}     &  \cmark  &      0.279          &    1587.10      &    0.808    \\
        \hline
        PAD3R (w/ static camera) &  \xmark & \cellcolor{tabfirst}0.126      &  \cellcolor{tabfirst}613.73      &    \cellcolor{tabsecond}0.941       \\
        PAD3R & \cmark & 0.137       &  \cellcolor{tabsecond}645.09      &    \cellcolor{tabfirst}0.942       \\
        \bottomrule
    \end{tabular}%
    }
    \Description{Table with quantitative Results on Consistent4D Dataset.}
    \label{tab:quant_c4d}
\end{table}

%% file: tables/quantative_comp.tex
\begin{table}[t]
    \centering
    \caption{\textbf{Quantitative Results on Artemis Dataset.} 
    The Artemis Dataset consists of rendered turntable videos of synthetic animals covering different angles of 40 degrees and 140 degrees. PAD3R achieves the best result, as methods like STAG4D~\cite{zeng2025stag4d}, L4GM~\cite{ren2024l4gm}, and DreamMesh4D (DM4D) ~\cite{li2024dreammesh4d} were unable to handle large camera motion with a static camera assumption. $\dagger$ Note that we split the input videos into chunks for L4GM due to limited vram. $\colorbox[RGB]{255,179,179}{\:}$: best, $\colorbox[RGB]{255,217,179}{\:}$: second-best, $\colorbox[RGB]{255,255,179}{\:}$: third-best
}
    \resizebox{0.48\textwidth}{!}{%
    \begin{tabular}{lcccc}
        \toprule
        \thead{\textbf{Methods}} & \thead{\textbf{Camera} \\ \textbf{modeling}} & \thead{\textbf{LPIPS$~\downarrow$}} & \thead{\textbf{FVD$~\downarrow$}} & \thead{\textbf{CLIP $~\uparrow$}} \\
        \midrule
        STAG4D~\cite{zeng2025stag4d}  &\xmark     & 0.240   & 1512.97 & 0.862 \\
        L4GM$\dagger$ ~\cite{ren2024l4gm} &   \xmark     &   \cellcolor{tabthird}0.235 &  \cellcolor{tabsecond} 1397.53 & 0.888  \\
        DM4D~\cite{li2024dreammesh4d} &\xmark  & \cellcolor{tabsecond} 0.232 & 3020.65   & 0.904\cellcolor{tabthird}  \\

        BANMo~\cite{yang2022banmo}  &  \cmark  &   0.265 & 1682.29  &  0.837 \\
        \midrule
        PAD3R (w/ static camera)&  \xmark    &   0.257   & \cellcolor{tabthird}1473.05   & 0.909\cellcolor{tabsecond}   \\
        PAD3R &  \cmark  & \cellcolor{tabfirst}0.184   & \cellcolor{tabfirst}1060.04  & \cellcolor{tabfirst}0.930  \\
        \bottomrule
    \end{tabular}%
    }
    \Description{Table with quantitative Results on Artemis Dataset.}
    \label{tab:quant_artmeis}
\end{table}

%% file: figures/fig_ablation.tex
\begin{figure}[t]    
    \centering
    \includegraphics[trim={2.8cm 16.0cm 2cm 4.4cm},clip,width=\linewidth]{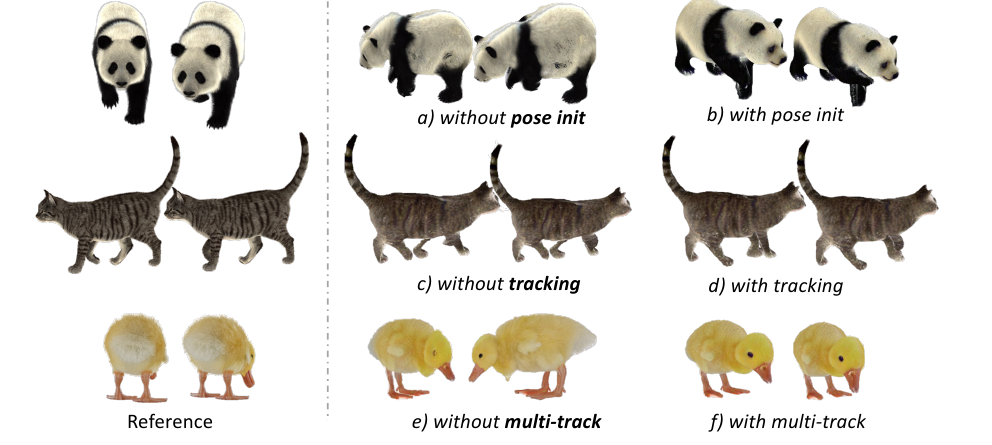}
    \vspace{-2mm}
    \caption{\textbf{Ablation.} We show the contributions of each of our components: a) without pose initialization vs b) with pose initialization, c) without tracking vs d) with tracking regularization, and e) without multi-tracking vs f) with multi-tracking regularization. We observe that without pose initialization in (a), there is a degradation in the panda's geometry and wrong orientation. Without tracking in (c), the reconstructed cat has an extra front leg. For (e), the orientation of the duck is wrong. Images sourced from Artemis (\textcopyright Luo et al., 2022).}
    \Description{Ablation study.}
    \label{fig:ablation}
\end{figure}

%% file: tables/quantative_comp_ablat.tex
\begin{table}[t]
    \centering
    \caption{\textbf{Ablation results on the Artemis dataset removing each component.} \textit{cam} stands for dynamic camera modeling. $\mathcal{P}_\mathrm{init}$ stands for initiating camera pose from PoseNet prediction. $\colorbox[RGB]{255,179,179}{\:}$: best, $\colorbox[RGB]{255,217,179}{\:}$: second-best, $\colorbox[RGB]{255,255,179}{\:}$: third-best.   
    }
    \begin{tabular}{lccc}
        \toprule
        \textbf{Method} & \textbf{LPIPS $\downarrow$} & \textbf{FVD $\downarrow$} &  \textbf{CLIP$\uparrow$} \\
        \midrule 
        cam   & 0.198 & \cellcolor{tabthird}1022.39 & 0.916 \\
        cam + $\mathcal{P}_{\mathrm{init}}$  & \cellcolor{tabthird}0.186 & 1048.95 & \cellcolor{tabsecond}0.927 \\ 
        cam + $\mathcal{P}_{\mathrm{init}}$ + $\mathcal{L}_{\text{track}}$ &   \cellcolor{tabsecond}0.181          &    \cellcolor{tabsecond}1015.07    &  \cellcolor{tabthird}0.926        \\
        full method (w. $\mathcal{L}_{\text{multi-track}}$) &     \cellcolor{tabfirst}{0.173}           &   \cellcolor{tabfirst}{1011.45}       &    \cellcolor{tabfirst}{0.928}       \\
        \bottomrule
    \end{tabular}%
    \Description{Table with ablation results on the Artemis dataset.}
    \label{tab:quant_ablt}
\end{table}

%% file: tables/quantative_alt_pose.tex
\begin{table}[t]
    \centering
    \caption{\textbf{Ablation results with off-the-shelf camera pose estimators.} 
    We compare our personalized PoseNet against state-of-the-art methods MegaSaM~\cite{megasam} and VGGT~\cite{vggt} on the Artemis dataset, evaluating reconstruction quality with and without camera refinement. Our PoseNet consistently achieves the best performance across all metrics.
    $\colorbox[RGB]{255,179,179}{\:}$: best, $\colorbox[RGB]{255,217,179}{\:}$: second-best.
}
    \resizebox{0.48\textwidth}{!}{%
    \begin{tabular}{lcccc}
        \toprule
        \thead{\textbf{Methods}} & \thead{\textbf{Camera} \\ \textbf{modeling}} & \thead{\textbf{LPIPS$~\downarrow$}} & \thead{\textbf{FVD$~\downarrow$}} & \thead{\textbf{CLIP $~\uparrow$}} \\
        \midrule
        MegaSaM~\cite{megasam}&\xmark &   0.214    &   1781.95   & 0.710     \\    
        VGGT~\cite{vggt} &\xmark&  \cellcolor{tabsecond}0.210   &  \cellcolor{tabsecond}1753.32    &  \cellcolor{tabsecond}0.729    \\
        PAD3R (PoseNet)&\xmark &     \cellcolor{tabfirst}{0.190} &   \cellcolor{tabfirst}{1180.52}  &   \cellcolor{tabfirst}{0.897}     \\
        \hline
        MegaSaM~\cite{megasam}& \cmark &   0.194    &   1094.42   & 0.833     \\    
        VGGT~\cite{vggt}& \cmark &   \cellcolor{tabsecond}0.174   &  \cellcolor{tabsecond}1026.94    &  \cellcolor{tabfirst}0.928    \\
        PAD3R (PoseNet)& \cmark &     \cellcolor{tabfirst}{0.173} &   \cellcolor{tabfirst}{1011.45}  &   \cellcolor{tabfirst}{0.928}     \\
        \bottomrule
    \end{tabular}
    }
    \vspace{-1em}
    \Description{Table compares with off-the-shelf camera pose estimators.}
    \label{tab:quant_alt}
\end{table}

%% file: 5_conclusions.tex
\section{Conclusion}
\label{sec:conclusions}

In this work, we present PAD3R for reconstructing dynamic, articulated objects from casually captured monocular videos. We address the limitations of both optimization-based methods and recent learning-based approaches. 
We achieve high-fidelity reconstructions that preserve geometric and texture integrity by leveraging an instance-specific pose estimator, a generative prior, and a novel multi-tracking regularization. 
Our evaluation of the Artemis video dataset highlights the improvements of our method over existing baselines, particularly in handling videos with large camera and object root-body motion. 
Our work provides a robust and flexible framework for extracting dynamic 3D representations from casual videos, thereby broadening the applicability of 4D reconstruction techniques to various real-world scenarios.

%% file: 6_acknowledgment.tex
\topic{Acknowledgments.}
We sincerely appreciate Shunsuke Saito for his early explorations with us, insightful discussions, and generous support.